\documentclass[conference]{IEEEtran}
\IEEEoverridecommandlockouts

\usepackage{cite}
\usepackage{amsmath,amssymb,amsfonts}
\usepackage{algorithmic}
\usepackage{graphicx}
\usepackage{textcomp}
\usepackage{xcolor}
\usepackage{booktabs}
\usepackage{hyperref}
\usepackage[numbers]{natbib}

\def\BibTeX{{\rm B\kern-.05em{\sc i\kern-.025em b}\kern-.08em
    T\kern-.1667em\lower.7ex\hbox{E}\kern-.125emX}}

\usepackage{comment}
\usepackage{academicons}
\definecolor{orcidlogocol}{HTML}{A6CE39}
\def\BibTeX{{\rm B\kern-.05em{\sc i\kern-.025em b}\kern-.08em
    T\kern-.1667em\lower.7ex\hbox{E}\kern-.125emX}}

\usepackage{scalerel}
\usepackage{tikz}
\usetikzlibrary{svg.path}

\definecolor{orcidlogocol}{HTML}{A6CE39}
\tikzset{
  orcidlogo/.pic={
    \fill[orcidlogocol] svg{M256,128c0,70.7-57.3,128-128,128C57.3,256,0,198.7,0,128C0,57.3,57.3,0,128,0C198.7,0,256,57.3,256,128z};
    \fill[white] svg{M86.3,186.2H70.9V79.1h15.4v48.4V186.2z}
                 svg{M108.9,79.1h41.6c39.6,0,57,28.3,57,53.6c0,27.5-21.5,53.6-56.8,53.6h-41.8V79.1z M124.3,172.4h24.5c34.9,0,42.9-26.5,42.9-39.7c0-21.5-13.7-39.7-43.7-39.7h-23.7V172.4z}
                 svg{M88.7,56.8c0,5.5-4.5,10.1-10.1,10.1c-5.6,0-10.1-4.6-10.1-10.1c0-5.6,4.5-10.1,10.1-10.1C84.2,46.7,88.7,51.3,88.7,56.8z};
  }
}

\newcommand\orcidicon[1]{\href{https://orcid.org/#1}{\mbox{\scalerel*{
\begin{tikzpicture}[yscale=-1,transform shape]
\pic{orcidlogo};
\end{tikzpicture}
}{|}}}}

\begin{document}

\title{FLARE-BO: Fused Luminance and Adaptive Retinex Enhancement via Bayesian Optimisation for Low-Light Robotic Vision}


\author{Nathan Shankar$^{\orcidicon{0000-0002-7255-2985}}$, Pawel Ladosz$^{\orcidicon{0000-0002-1154-8333}}$ and Hujun Yin$^{\orcidicon{0000-0002-9198-5401}}$ 
\thanks{This work was supported by the Robotics and Artificial Intelligence Collaboration (RAICo).
(Corresponding author: Pawel Ladosz.)
The authors are with the Department of Mechanical, Aerospace and Civil Engineering and the Department of Electrical and Electronic Engineering at the University of Manchester, Manchester, M13 9PL, United Kingdom, (e-mail: pawel.ladosz@manchester.ac.uk).
}}

\maketitle

\begin{abstract}
Reliable visual perception under low illumination remains a core
challenge for autonomous robotic systems, where degraded image quality
directly compromises navigation, inspection, and various operations.
A recent training free approach 
showed that Bayesian optimisation with Gaussian Processes can
adaptively select brightness, contrast, and denoising parameters on a per-image basis, achieving competitive enhancement without any learned model.
However, that framework is limited to three parameters, applies no
illumination decomposition or white balance correction, and relies on
Non-Local Means denoising, which tends to over smooth edges under noisy conditions.
This paper proposes FLARE-BO (Fused
Luminance and Adaptive Retinex
Enhancement via Bayesian Optimisation),
an extended framework that jointly optimises eight parameters spanning across
gamma correction, LIME-style illumination normalisation, chrominance
denoising, bilateral filtering, NLM denoising, Grey-World automatic
white balance, and adaptive post smoothing.
The search engine employs a unit hypercube parameter normalisation,
objective standardisation, Sobol quasi-random initialisation, and Log
Expected Improvement acquisition for principled exploration of the
expanded space.
Performance of the proposed method is benchmarked using the Low Light paired dataset (LOL) and results show marked improvements of the proposed method over existing methods that were not specifically trained using this dataset.
\end{abstract}


\section{Introduction}

Autonomous robotic systems operating in uncontrolled environments must
maintain reliable visual perception across a wide range of illumination
conditions.
Applications including subsea navigation~\cite{wu2019ultra}, structural
inspection ~\cite{schwaiger2024ugv}, and search-and-rescue
missions~\cite{abdeh2021autonomous} which routinely encounter scenes where the
ambient light falls far below the threshold required for standard
camera operation.
Under such conditions, sensor noise is amplified, local contrast
within the image degrades, colour channels are unevenly degraded, and fine spatial
detail is lost which in turn impairs the downstream perception
modules that robotic systems depend on.

Rodrigues~\textit{et al.}~\cite{rodrigues2025low} proposed a training-free
alternative by casting enhancement as a black-box parameter optimisation
problem.
Their method applies a linear brightness and contrast transform followed
by Non-Local Means (NLM) denoising, and uses a Gaussian Process (GP)
surrogate to search for the optimal three-parameter configuration
$(\alpha, \beta, h)$ that maximises a composite quality objective on
each image individually.
On the LOL dataset~\cite{wei2018deep} it achieved PSNR~18.79,
SSIM~0.711, and NIQE~3.909, outperforming several deep learning
methods on structural fidelity without any training data.

While this represents a compelling baseline, the framework has
several limitations that constrain its enhancement ceiling.
The three-parameter pipeline omits illumination decomposition entirely,
meaning that spatially non uniform darkness, the dominant mode in
real low-light scenes, is handled only through a global linear
rescaling.
No chrominance correction is applied, leaving colour noise and
channel-level bias unaddressed.
NLM denoising treats all spatially distant patches equally within its
search window, which tends to over smooth edges in signal dependent
noise regions.
Finally, the unscaled and unstandardised optimisation space can
bias GP kernel learning, and standard expected improvement may
under explore in later iterations.

This paper addresses these limitations with \textbf{FLARE-BO}, an
eight parameter Bayesian optimisation framework for low-light
enhancement.
The key contributions are:
\begin{enumerate}
    \item A structured eight-stage enhancement pipeline incorporating
          LIME style guided illumination normalisation, gamma and
          linear contrast correction, Grey World automatic white
          balance (AWB), LAB space chrominance denoising, bilateral
          spatial filtering, NLM luminance denoising, and adaptive
          Gaussian post-smoothing which of each are controlled by a dedicated
          optimised parameter.
    \item A structure prioritised composite objective function
           $f(\theta)$ that strongly emphasises structural
          similarity, steering the search toward perceptually coherent
          outputs suited to downstream robotic perception.
    \item A numerically robust BO engine featuring unit hypercube
          parameter normalisation, objective standardisation,
          Sobol initialisation, and LogEI acquisition with
          L-BFGS-B maximisation.
\end{enumerate}

The remainder of the paper is structured as follows.
Section~\ref{sec:related} reviews related work.
Section~\ref{sec:method} presents the FLARE-BO pipeline and
optimisation engine.
Section~\ref{sec:experiments} reports experimental results.
Section~\ref{sec:conclusions} concludes the findings of this work and suggest future directions.

\section{Related Work}
\label{sec:related}

\subsection{Classical and Heuristic Methods}

Early low-light enhancement methods were based of off Retinex theory,
which separates a scene into an illumination layer and a reflectance
layer~\cite{pham2020low}.
Global operators such as histogram equalisation and contrast stretching
improved the average brightness but applied uniform transformations that
failed under spatially varying illumination.
LIME~\cite{guo2016lime} improved upon global Retinex by constructing
a spatially varying illumination map through a structure aware
regularisation but its fixed smoothing parameters limit
adaptability across different scene types.
Gamma correction~\cite{kubinger1998role} and multiscale Retinex variants~\cite{lee2013adaptive}
offered controllable lifting but required manual parameter selection
for each dataset.
None of these approaches jointly addressed illumination normalisation,
colour correction, noise suppression, and structural preservation
under a unified adaptive framework.

\subsection{Deep Learning Methods}

Supervised deep networks have substantially raised the enhancement
standards and benchmark.
RetinexNet~\cite{wei2018retinexnet} decomposed images into reflectance
and illumination components using paired training data.
Zero-DCE~\cite{guo2020zerodce} reformulated enhancement as
zero reference curve estimation guided by non-reference losses for
colour constancy, smoothness, and illumination range.
EnlightGAN~\cite{jiang2021enlightgan} used unpaired adversarial
training to avoid the requirement for reference images.
MIRNet~\cite{zamir2020learning} combined multiscale feature aggregation
with selective kernel fusion for joint denoising and detail recovery.
URetinex-Net~\cite{wu2022uretinexnet} embedded Retinex priors into a
deep unfolding network, while SNR-aware enhancement~\cite{xu2022snr}
modulated processing according to local signal-to-noise estimates.
RetinexFormer~\cite{cai2023retinexformer} brought one-stage transformer
attention into the Retinex framework, and PyDiff~\cite{zhou2023pyramid}
showed that pyramid diffusion models could achieve high perceptual
quality, though at substantial computational cost.
The shared limitation of all these approaches is their dependence on
large paired datasets.

\subsection{Optimisation Based Methods}

RUAS~\cite{liu2021ruas} reduced data dependence by unrolling an energy
minimisation with learnable activation functions, but still required a
training phase.
Semantic aware guidance~\cite{wu2023semantic} incorporated scene level
priors into image restoration but again relied on supervised learning.

Rodrigues~\textit{et al.}~\cite{rodrigues2025low} proposed a Bayesian optimisation approach to low-light
enhancement, using a GP surrogate to select per-image brightness,
contrast, and NLM denoising parameters guided by a PSNR, SSIM and NIQE
composite objective.
The approach is immediately applicable to any sensor without data
collection, making it well-suited for a diverse range of low-light datasets without prior training, however it is failed to note that the optimisation itself is a a time consuming process and cannot be used for online denoising, and their method is only practical for offline denoising.
FLARE-BO builds directly on this work, substantially expanding the
parameter space, introducing illumination decomposition and chrominance
correction stages, and improving the optimisation engine, yielding
markedly higher quantitative performance on the same benchmark.

\section{Proposed Method}
\label{sec:method}

\begin{figure*}[t]
    \centering
    \includegraphics[width=1.0\textwidth]{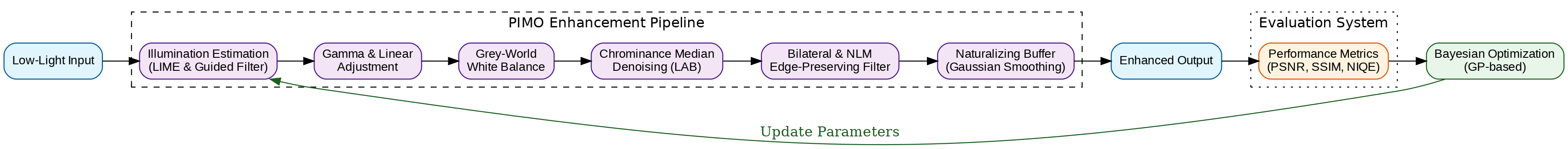}
    \caption{Architecture of the enhancement technique}
    \label{fig:pipeline}
\end{figure*}

FLARE-BO processes each low-light image $I$ through a structured
eight-stage pipeline whose parameters
$\boldsymbol{\theta} = (\alpha, \beta, \gamma, h, \sigma_s,
\lambda, d, h_c)$ are selected per image by Bayesian Optimisation.
\autoref{fig:pipeline} illustrates the architecture.
Table~\ref{tab:params} summarises all parameters and their search bounds.

\begin{table}[!htbp]
\centering
\caption{FLARE-BO optimised parameters and search bounds.}
\label{tab:params}
\begin{tabular}{lllc}
\toprule
\textbf{Symbol} & \textbf{Description} & \textbf{Stage} & \textbf{Bounds} \\
\midrule
$\alpha$   & Contrast scale factor       & Linear adjust  & $[0.5,\;5.0]$ \\
$\beta$    & Brightness offset           & Linear adjust  & $[-20,\;50]$ \\
$\gamma$   & Gamma exponent              & Gamma correct  & $[0.1,\;2.0]$ \\
$\lambda$  & LIME illumination factor    & Illumination   & $[0.0,\;0.6]$ \\
$h_c$      & Chrominance filter strength & Chroma denoise & $[0,\;40]$ \\
$d$        & Bilateral filter diameter   & Spatial filter & $[0,\;15]$ \\
$h$        & NLM luminance strength      & NLM denoise    & $[1,\;50]$ \\
$\sigma_s$ & Post-smoothing std.\ dev.   & Smoothing      & $[0.0,\;1.5]$ \\
\bottomrule
\end{tabular}
\end{table}

\subsection{Stage 1: LIME-Style Illumination Normalisation}

When $\lambda > 0$, a Retinex-inspired illumination correction is
applied first.
The per-pixel illumination map is estimated as the maximum channel
response:
\begin{equation}
    L(x) = \max_{c \in \{R,G,B\}} I_c(x)
    \label{eq:ilum}
\end{equation}
and smoothed with a guided filter~\cite{he2013guided} to produce an
edge-preserving illumination estimate $\hat{L}(x)$.
The image is then normalised:
\begin{equation}
    I_\mathrm{lime}(x) =
        \frac{I(x)}{\hat{L}(x)^{\lambda} + \epsilon}
    \label{eq:lime}
\end{equation}
where $\epsilon = 10^{-4}$ prevents division by zero and $\lambda$
controls the strength of the correction.
Setting $\lambda = 0$ bypasses this stage entirely.

\subsection{Stage 2: Gamma and Linear Contrast Adjustment}

A combined gamma correction and linear rescaling is applied:
\begin{equation}
    I' = \mathrm{clip}\!\bigl(\alpha \cdot I_\mathrm{lime}^{\gamma}
         \cdot 255 + \beta,\; 0,\; 255\bigr)
    \label{eq:gamma}
\end{equation}
where $\alpha$ is the linear contrast factor,
$\gamma$ is the gamma exponent, and $\beta$ is the
additive brightness offset.
The exponentiation is applied in the normalised $[0,1]$ domain before
rescaling to 8-bit.

\subsection{Stage 3: Grey World Automatic White Balance}

After contrast adjustment, a Grey World AWB
correction~\cite{buchsbaum1980spatial} is applied to $I'$.
The method assumes that the mean of each colour channel should equal
the global mean intensity $\bar{\mu}$ under neutral illumination.
Per channel gain factors are computed as:
\begin{equation}
    g_c = \frac{\bar{\mu}}{\mu_c + \epsilon},
    \quad c \in \{R,G,B\}
    \label{eq:awb}
\end{equation}
and each channel is scaled and clamped accordingly.
This correction is parameter free and it introduces no additional
degree of freedom to the search besides it removes the chromatic bias
that low light captures consistently exhibit under artificial
illumination sources.

\subsection{Stage 4: LAB Space Chrominance Denoising}

Sensor noise under low illumination manifests as both luminance
noise and chrominance noise. The latter produces visually disturbing random colour speckle that is
not addressed by luminance only denoisers.
When $h_c > 0$, FLARE-BO converts $I'$ to the CIELAB colour space
and applies a median filter separately to the $a^*$ and $b^*$
chrominance channels:
\begin{equation}
    k = 2\left\lfloor \frac{h_c}{10} \right\rfloor + 1
    \label{eq:ksize}
\end{equation}
where $k \geq 3$ is the filter kernel size.
The median filter is chosen over Gaussian smoothing because it
suppresses impulse type colour noise without blurring
chrominance edges.
The luminance channel $L^*$ is left unmodified at this stage,
preserving brightness information for the subsequent denoising steps.

\subsection{Stage 5: Bilateral Spatial Filtering}

A bilateral filter is applied to the chrominance corrected image
when $d > 0$:
\begin{equation}
\tilde{I}(x) = \frac{1}{W(x)}
\sum_{y \in \mathcal{N}_d(x)}
G_s\!\bigl(\|x - y\|\bigr)\,
G_r\!\bigl(\|\tilde{I}(x) - \tilde{I}(y)\|\bigr)\,
\tilde{I}(y)
\label{eq:bilateral}
\end{equation}
\begin{equation}
\mathcal{N}_d(x) = \{ y \mid \|x - y\| \leq d/2 \}
\label{eq:neighbourhood}
\end{equation}
where $G_s$ and $G_r$ are spatial and radiometric Gaussian kernels
respectively, $W(x)$ is the per-pixel normalisation constant, and $d$ is the filter diameter.
Fixed sigma values are used for both kernels, with $d$ as the
sole optimised parameter.
The joint spatial radiometric weighting preserves intensity
discontinuities at object boundaries while suppressing noise in
smooth regions, providing more better edge smoothing than NLM
for signal-dependent noise~\cite{tomasi1998bilateral}.

\subsection{Stage 6: NLM Luminance Denoising}

After bilateral filtering, NLM denoising~\cite{buades2005nlm} is
applied with luminance filter strength $h$.
NLM computes a weighted average over non-local similar patches:
\begin{equation}
    I''(x) = \sum_{y \in \mathcal{N}(x)} w(x,y)\,\tilde{I}(y)
    \label{eq:nlm}
\end{equation}
\begin{equation}
    w(x,y) \propto \exp\!\bigl(-\|\tilde{I}(x) - \tilde{I}(y)\|^2 /h^2\bigr)
\end{equation}

Following bilateral filtering, NLM suppresses the residual fine grained luminance noise
that bilateral filtering may miss in the high-frequency texture regions.
The two denoising stages are therefore interdependent, where the bilateral
filtering handles large scale smooth regions and edges, while the NLM
handles fine grained textural noise.
The colour denoising strength is set to $h + h_c$ to maintain
consistency with the chrominance correction applied in Stage~4.

\subsection{Stage 7: Adaptive Gaussian Post-Smoothing}

A final Gaussian smoothing pass with standard deviation
$\sigma_s$ is applied to suppress quantisation artefacts.
When $\sigma_s \leq 0.05$ the pass is skipped entirely.
As $\sigma_s$ increases, fine texture detail is traded for smoother
tonal transitions, which tends to benefit NIQE at the cost of some
SSIM.
This parameter allows the optimiser to explicitly control the
perceptual naturalness of the final output.

\subsection{Bayesian Optimisation Engine}
\label{sec:bo}

\begin{figure}
    \centering
    \includegraphics[width=1\linewidth]{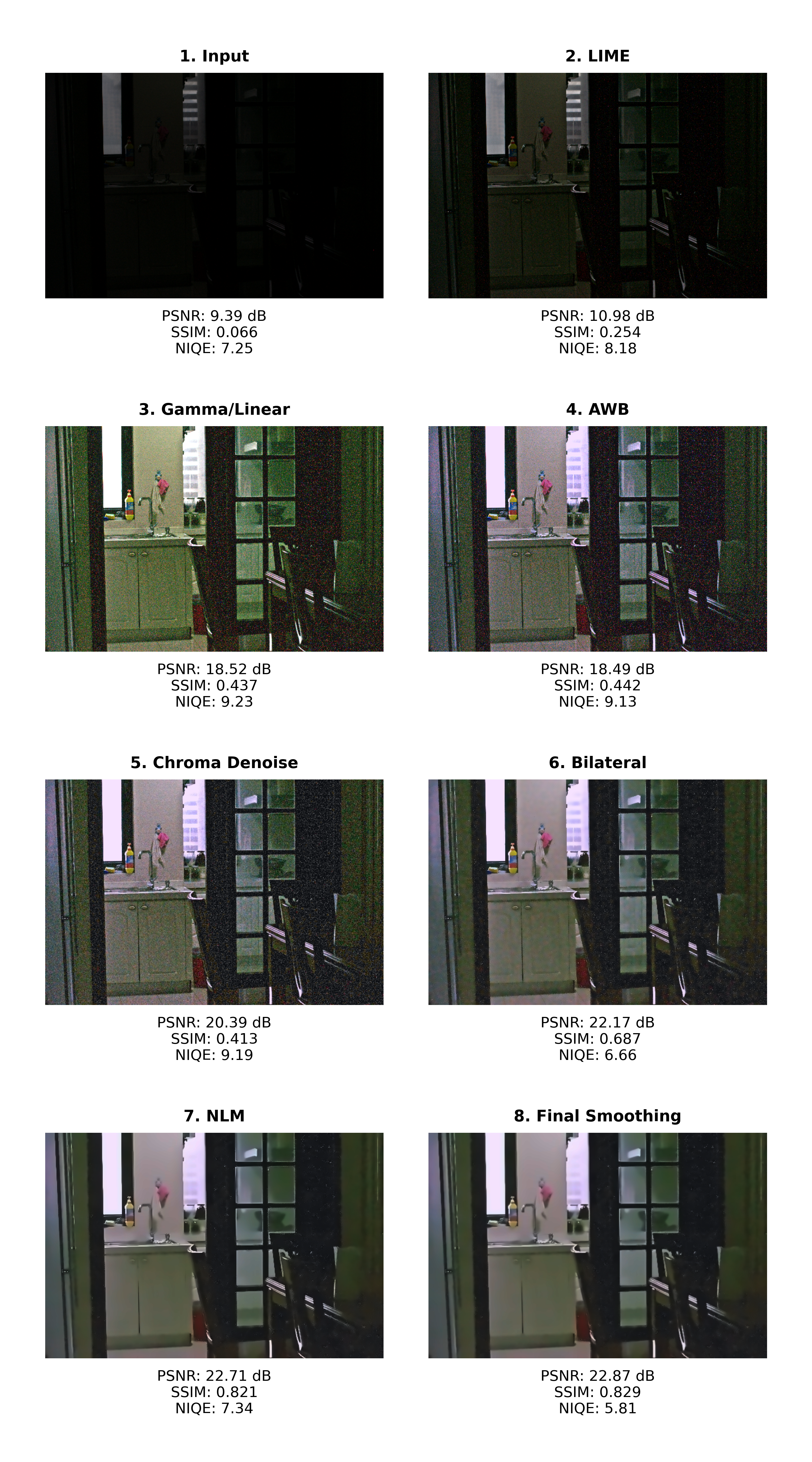}
    \caption{The different stages of enhancement in the pipeline}
    \label{fig:stages}
\end{figure}
All eight parameters
$\boldsymbol{\theta} = (\alpha, \beta, \gamma, h, \sigma_s, \lambda,
d, h_c)$ are optimised jointly by Bayesian Optimisation with a GP
surrogate~\cite{snoek2012practical}.
The scalar objective is:
\begin{equation}
    f(\boldsymbol{\theta}) =
        1.0\,\mathrm{PSNR}
        + 80.0\,\mathrm{SSIM}
        - 5.0\,\mathrm{NIQE}
    \label{eq:objective}
\end{equation}
where the coefficients are set empirically to compensate for the
differing numerical scales of each metric as PSNR typically ranges
from 10-25~dB on LOL, SSIM from 0.4-0.9, and NIQE from 3-10.
The weight of 80.0 on SSIM counteracts its compressed $[0,1]$ range
while deliberately prioritising structural fidelity and making it just as important.
The coefficient of 5.0 on NIQE treats perceptual naturalness as a
secondary regulariser rather than a primary driver.
PSNR is always numerically larger regardless of the dataset, the relative weighting remains meaningful beyond LOL and sensitivity analysis across additional
benchmarks is left as future work.

\textbf{Parameter normalisation:}
All raw parameters are mapped to the unit hypercube $[0,1]^8$ before
any GP operation:
\begin{equation}
    \hat{\theta}_i =
        \frac{\theta_i - \theta_i^{\min}}
             {\theta_i^{\max} - \theta_i^{\min}}
    \label{eq:scale}
\end{equation}
This removes the large scale disparity between parameters
like $\alpha$ and $h$, ensuring that
the GP squared exponential kernel assigns equitable length scale
learning to all dimensions.
Sobol sampling, acquisition maximisation, and all GP fitting are
performed in the unit cube and the candidates are unscaled to raw bounds
before evaluation.

\textbf{Objective standardisation:}
Collected objective values are standardised before GP model fitting:
\begin{equation}
    \tilde{f}_i = \frac{f_i - \bar{f}}{\sigma_f + \epsilon}
    \label{eq:standardise}
\end{equation}
where $\bar{f}$ and $\sigma_f$ are the running mean and standard
deviation, and $\epsilon = 10^{-6}$ prevents degenerate scaling.
Standardisation stabilises GP kernel hyperparameter optimisation
across the wide range of absolute objective values that arise across
different images.

\textbf{Initialisation:}
Before active optimisation, candidates are drawn from the
unit hypercube using Sobol quasi-random sampling~\cite{sobol1967}.
The larger initialisation set
reflects the expanded eight-dimensional search space, where
low discrepancy coverage is more critical for avoiding poor
warm start optima.

\textbf{Acquisition:}
Each subsequent candidate is selected by maximising the Log Expected
Improvement~\cite{ament2024unexpected}:
\begin{equation}
    \mathrm{LogEI}(\hat{\boldsymbol{\theta}})
        = \log\,\mathbb{E}\!\Bigl[
            \max\!\bigl(\tilde{f}(\hat{\boldsymbol{\theta}})
                        - \tilde{f}^*,\;0\bigr)
          \Bigr]
    \label{eq:logei}
\end{equation}
where $\tilde{f}^*$ is the best standardised objective observed so
far.
LogEI is numerically stable in flat regions of the acquisition
surface where standard EI can collapse to near-zero values,
maintaining meaningful candidate differentiation in later iterations.
The acquisition function is maximised via multi-start L-BFGS-B.

\textbf{Model fitting:}
The GP (SingleTaskGP with automatic kernel learning) is implemented
in BoTorch~\cite{balandat2020botorch} and GPyTorch~\cite{gardner2018gpytorch}
and refitted at every iteration by exact marginal log-likelihood
optimisation.
The engine is reinitialised independently for each input image,
ensuring fully image adaptive parameter selection.

The enhancement done across all the explained stages in the pipeline can be seen in \autoref{fig:stages}. This shows how the different stages improves the metrics of the image while improving the structure, reducing the noise and making the images look more natural.

\section{Experiments}
\label{sec:experiments}

\subsection{Dataset}

Experiments are conducted on the \textbf{LOL} (Low-Light paired)
dataset~\cite{wei2018deep}, which comprises 500 paired low-light and
normal-light images, all at $400 \times 600$
resolution depicting predominantly indoor scenes.

\begin{figure*}[t]
    \centering
    \includegraphics[width=0.70
    \linewidth]{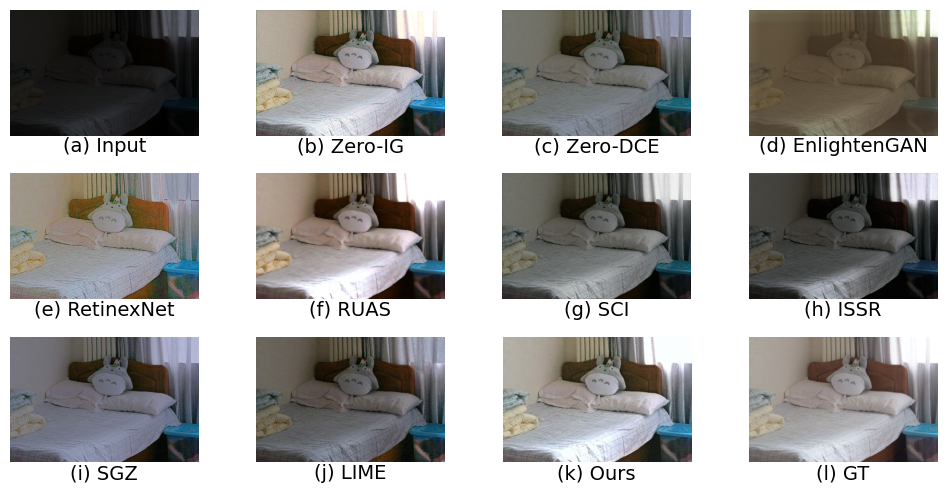}
    \caption{Comparison across various enhancement techniques.}
    \label{fig:comparison}
\end{figure*}

\subsection{Compared Methods}

FLARE-BO is evaluated against nine published methods as seen in \autoref{fig:comparison} spanning across
heuristic, deep learning, and optimisation-based categories.
The heuristic category includes LIME~\cite{guo2016lime} and
ZeroIG~\cite{shi2024zero}.
The deep learning category includes
Zero-DCE~\cite{guo2020zerodce},
EnlightenGAN~\cite{jiang2021enlightgan},
RetinexNet~\cite{wei2018retinexnet},
RUAS~\cite{liu2021ruas},
SCI~\cite{ma2022toward},
ISSR~\cite{fan2020integrating}, and
SGZ~\cite{zheng2022semantic}.
Methods with publicly available weights specifically trained on the LOL dataset are not included in this comparison, as they benefit from dataset specific supervision. In contrast, all evaluated methods operate without explicit training on LOL, ensuring a fair comparison with FLARE-BO, which is training free.
Rodrigues~\textit{et al.}~\cite{rodrigues2025low} is used as the primary baseline, as FLARE-BO directly extends its training free optimisation framework, and all methods are evaluated on the same LOL dataset. However, since the method in~\cite{rodrigues2025low} is not open-sourced, it could not be independently evaluated under the same conditions as other methods. Therefore, the results reported for this method are taken directly from the original paper.

\subsection{Evaluation Metrics}

Three metrics are used to assess enhancement quality with benchmarks to the previous work.
PSNR~\cite{hore2010image} measures pixel-level signal fidelity
between the enhanced output and the reference image; higher values
indicate closer agreement.
SSIM~\cite{wang2004image} jointly evaluates luminance, contrast, and
structural correlation, providing a perceptually motivated measure of
image similarity; higher is better.
NIQE~\cite{mittal2012making} is a no-reference metric that quantifies
deviation from the statistical properties of natural pristine images;
lower scores indicate higher perceptual naturalness.
Because LOL provides paired ground-truth references, PSNR and SSIM
are the primary indicators of reconstruction fidelity, with NIQE
serves as a supplementary naturalness check.

\subsection{Quantitative Results}

Table~\ref{tab:results} reports the quantitative comparison.
FLARE-BO achieves the highest PSNR and SSIM
of all compared methods by a substantial margin.
The closest competitor on PSNR and SSIM is Rodrigues~\textit{et
al.}~\cite{rodrigues2025low}.

The results reflect the cumulative benefit of the expanded
eight parameter pipeline over the three parameter baseline
of~\cite{rodrigues2025low}.
The LIME-style illumination normalisation corrects spatially varying
darkness before any global transform is applied, a capability absent
from linear-only pipelines. 
The gamma exponent introduces a non-linear tonal response that more
closely matches the human visual system's sensitivity under dark
adaptation.
The LAB-space chrominance denoising removes colour speckle that
luminance-domain filters such as bilateral and NLM denoisers cannot
address on their own.
Together these additions substantially improve structural
correspondence with the ground-truth reference.

On NIQE, Rodrigues~\textit{et al.}~\cite{rodrigues2025low} achieve
the best score and ISSR the second best.
FLARE-BO scores 4.695, which is higher than all other baselines
including RetinexNet, ZeroIG, SGZ, LIME, Zero-DCE, SCI, EnlightenGAN, and
RUAS.
The moderately higher NIQE relative to~\cite{rodrigues2025low} is
the expected consequence of the SSIM dominant weighting in
Equation~\eqref{eq:objective} as the optimiser is steered away from
the heavy global smoothing that lowers NIQE but undermines the structural
detail that SSIM rewards. Giving NIQE a higher coefficient made the optimiser struggle between balancing all three parameters and this caused the resulting image to struggle with a good PSNR and SSIM. 

\begin{table}[t]
\centering
\caption{Image quality assessment on the LOL dataset.
         $\uparrow$ higher is better; $\downarrow$ lower is better.
         \textbf{Bold}: best. \underline{Underline}: second best.}
\label{tab:results}
\begin{tabular}{lccc}
\toprule
\textbf{Method} & \textbf{PSNR}~$\uparrow$ & \textbf{SSIM}~$\uparrow$
                & \textbf{NIQE}~$\downarrow$ \\
\midrule
ZeroIG~\cite{shi2024zero}
    & 16.580 & 0.531 & 8.938 \\
Zero-DCE~\cite{guo2020zerodce}
    & 14.585 & 0.611 & 8.081 \\
EnlightenGAN~\cite{jiang2021enlightgan}
    & 13.604 & 0.659 & 7.026 \\
RetinexNet~\cite{wei2018retinexnet}
    & 17.120 & 0.498 & 9.300 \\
RUAS~\cite{liu2021ruas}
    & 18.206 & 0.715 & 6.867 \\
SCI~\cite{ma2022toward}
    & 13.609 & 0.544 & 7.246 \\
ISSR~\cite{fan2020integrating}
    & 12.442 & 0.544 & \underline{4.026} \\
SGZ~\cite{zheng2022semantic}
    & 15.238 & 0.620 & 8.249 \\
LIME~\cite{guo2016lime}
    & 13.745 & 0.571 & 8.294 \\
Rodrigues~\textit{et al.}~\cite{rodrigues2025low}
    & \underline{18.790} & \underline{0.711} & \textbf{3.909} \\
\midrule
\textbf{FLARE-BO (proposed)}
    & \textbf{22.402} & \textbf{0.8427} & 4.695 \\
\bottomrule
\end{tabular}
\end{table}

\subsection{Qualitative Results}

\begin{figure}[!h]
    \centering
    \includegraphics[width=1\linewidth]{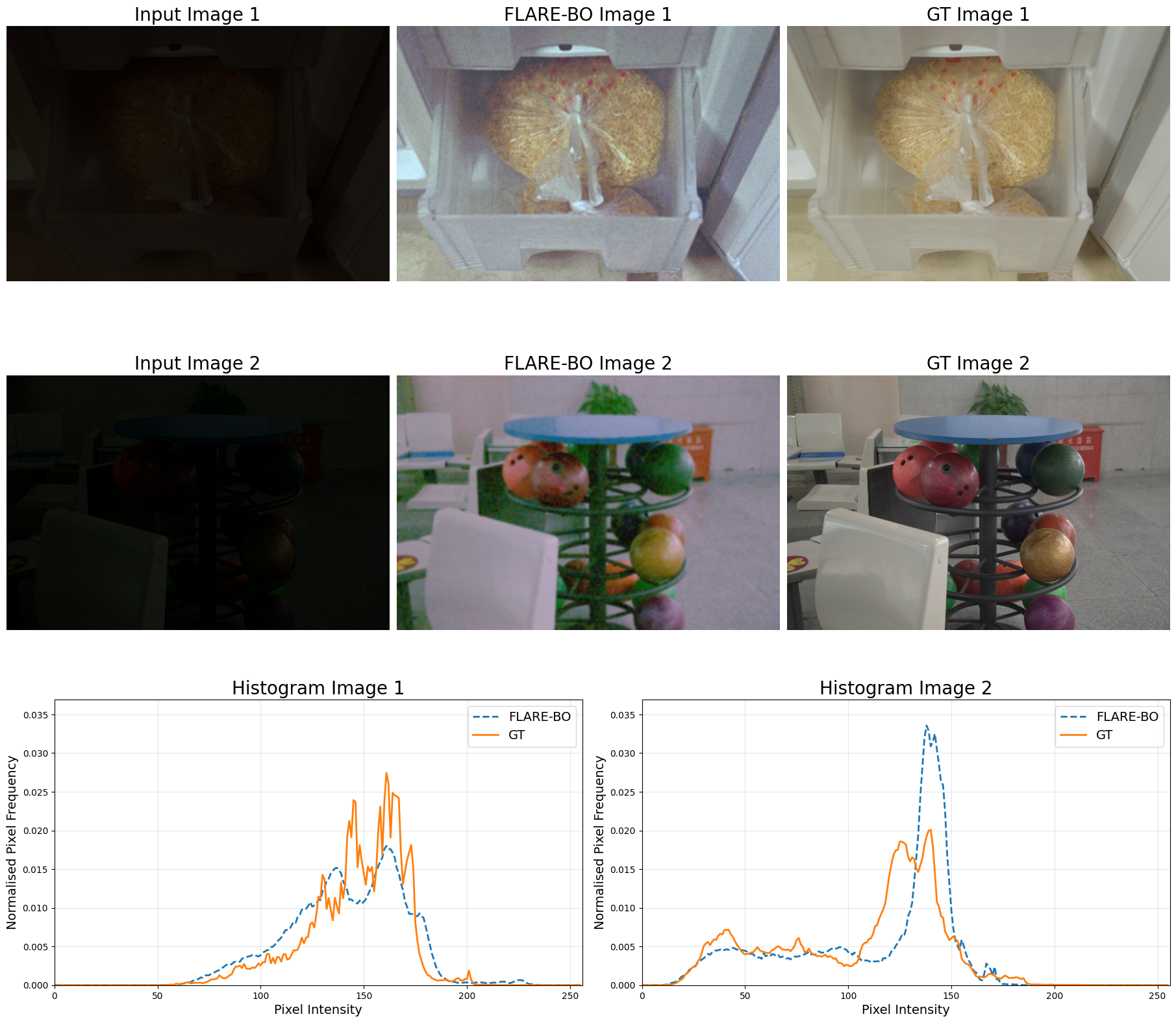}
    \caption{Sample Images from the LOL Dataset.}
    \label{fig:hist}
\end{figure}

Visual inspection from Fig.~\ref{fig:hist} confirms the quantitative findings. FLARE-BO produces images with consistent brightness across the full frame, neutral colour rendition, and well preserved structural detail. The illumination normalisation stage is particularly effective in scenes with mixed or spatially varying lighting, where the input contains both heavily underexposed regions and locally brighter objects. FLARE-BO restores visibility in both cases without introducing saturation artefacts, clipping, or tonal inversion

The histogram analysis further supports this behaviour. In both scenes, the input distributions are heavily skewed toward low intensities, reflecting severe underexposure. In contrast, FLARE-BO shifts the distribution toward the mid intensity range while maintaining a similar overall shape to the ground truth. This indicates that the enhancement is not a simple global brightness boost, but a structured remapping of intensities that preserves relative luminance relationships. Unlike competing methods that either over compress the dynamic range or introduce artificial peaks, FLARE-BO maintains a smoother, more continuous distribution that closely tracks the ground-truth statistics across all tonal regions. Chrominance denoising visibly eliminates the random colour speckle that persists in results in some compared methods. FLARE-BO jointly optimises noise suppression and structural fidelity under a single objective, resulting in outputs that are both perceptually stable and statistically aligned with ground truth.
\section{Conclusions}
\label{sec:conclusions}

This paper presented FLARE-BO, a training free low-light image
enhancement framework that  extends the Bayesian
optimisation of Rodrigues~\textit{et al.}~\cite{rodrigues2025low}.
By expanding from three to eight jointly optimised parameters and
adding LIME-style illumination normalisation, gamma correction,
Grey-World white balance, LAB-space chrominance denoising, bilateral
filtering, and adaptive post-smoothing, FLARE-BO achieves PSNR
improvement of $+3.61$~dB and SSIM improvement of $+0.132$ on the
LOL dataset relative to the baseline~\cite{rodrigues2025low}, while
outperforming all compared deep learning methods on both
reference based metrics.
The BO engine with unit hypercube normalisation,
objective standardisation, Sobol initialisation, and LogEI acquisition
ensures effective exploration of the expanded eight dimensional search
space.

The method requires no training data making it usable on any kind of low-light image without prior scene information. Its primary limitation is it has a higher NIQE, which is the deliberate consequence of heavily weighting SSIM in the objective function and also the optimisation process is slower than deep learning methods and this optimiser can only be run offline. Future work will explore incorporating perceptual loss terms to
recover NIQE competitiveness, multi-scale illumination estimation for
high resolution inputs, dynamic optimisation budget allocation based
on scene complexity, and evaluation on additional benchmarks

\section*{Acknowledgment}
This work was supported by the Robotics and Artificial Intelligence Collaboration (RAICo).

\bibliographystyle{IEEEtran}
\bibliography{references}

\end{document}